\documentclass[letterpaper, journal, twoside]{IEEEtran}
\IEEEoverridecommandlockouts

\usepackage{dsfont}
\usepackage[T1]{fontenc}
\usepackage{graphicx, stfloats}
\usepackage{geometry}
\usepackage{hyperref}
\usepackage{booktabs}
\usepackage{subcaption}
\usepackage{amssymb, amsmath}
\usepackage{algorithm}
\usepackage{algpseudocode}
\usepackage{makecell}
\usepackage{xcolor}
\usepackage{xurl}
\begin{document}
\markboth{Preprint Version. 2026}
{Khatim \& Arief (2026): Optimized Human-Robot Co-Dispatch 
Planning for Petro-Site Surveillance under Varying Criticalities}

\title{Optimized Human-Robot Co-Dispatch Planning for Petro-Site Surveillance under Varying Criticalities}

\author{\IEEEauthorblockN{Nur Ahmad Khatim}
\IEEEauthorblockA{\textit{Institut Teknologi Sepuluh Nopember (ITS)} \\
\textit{Informatics Engineering}\\
Surabaya, Indonesia \\
\orcidlogo \enspace \href{https://orcid.org/0009-0008-6939-8121}{\color{blue}\uline{https://orcid.org/0009-0008-6939-8121}}
}
\and
\IEEEauthorblockN{Mansur M. Arief}
\IEEEauthorblockA{\textit{King Fahd University of Petroleum and Minerals (KFUPM)} \\
\textit{Industrial and Systems Engineering}\\
Dhahran, Saudi Arabia }
}

\author{Nur Ahmad Khatim$^{1}$, Mansur M. Arief$^{2\star}$%
\thanks{$^1$Nur Ahmad Khatim is with the Institut Teknologi 
Sepuluh Nopember (ITS), Surabaya, Indonesia}%
\thanks{$^2$Mansur M. Arief is with the Department of Industrial 
and Systems Engineering, Interdisciplinary
Research Center for Smart Mobility and
Logistics (IRC-SML), and KFUPM-SDAIA Joint
Research Center for AI (JRCAI), King Fahd University of Petroleum and 
Minerals (KFUPM), Dhahran, Saudi Arabia (\url{mansur.arief@kfupm.edu.sa)}}%
\thanks{$^{\star}$Corresponding Author}%
\thanks{Code available at: 
\url{https://github.com/ai-vnv/Petro-HRCD-FLP}}%
}

\maketitle

\begin{abstract}
Securing petroleum infrastructure requires balancing autonomous system efficiency with human judgment for threat escalation—a challenge unaddressed by classical facility location models assuming homogeneous resources. This paper formulates the Human-Robot Co-Dispatch Facility Location Problem (HRCD-FLP), a capacitated facility location variant incorporating tiered infrastructure criticality, human-robot supervision ratio constraints, and minimum utilization requirements. We evaluate command center selection across three technology maturity scenarios. Results show transitioning from conservative (1:3 human-robot supervision) to future autonomous operations (1:10) yields significant cost reduction while maintaining complete critical infrastructure coverage. For small problems, exact methods dominate in both cost and computation time; for larger problems, the proposed heuristic achieves feasible solutions in under 3 minutes with approximately 14\% optimality gap where comparison is possible. From systems perspective, our work demonstrate that optimized planning for human-robot teaming is key to achieve both cost-effective and mission-reliable deployments.
\end{abstract}

\begin{IEEEkeywords}
Human-robot teaming, facility location problem, critical infrastructure protection, systems security
\end{IEEEkeywords}

\section{Introduction}

\IEEEPARstart{T}{he} global petroleum infrastructure represents one of the most extensive critical infrastructure networks, with over 2.7 million miles of pipelines in the United States alone and similar networks spanning the Arabian Peninsula and other major hydrocarbon basins \cite{coburn2020oil, klass2014transporting, dietl2021global}. These networks exhibit a hierarchical topology: high-criticality nodes (refineries, processing facilities, storage terminals) interconnected by extensive linear assets of varying criticality. Protection of such infrastructure has emerged as a paramount concern for national security and economic stability \cite{pashchenko2024main, bajpai2007securing}, particularly as threat vectors now encompass cyber-physical attacks \cite{mohammed2022cybersecurity, sophos2024ransomware}.

The advent of the fourth industrial revolution technologies has catalyzed a paradigm shift in industrial security operations. Unmanned aerial vehicles, autonomous ground vehicles, and integrated sensor networks (collectively called \textit{robots} in this work) now complement human surveillance capacity. However, integrating these heterogeneous assets presents a complex systems engineering challenge---not merely technological, but fundamentally interdisciplinary among resource allocation, facility location, and human-machine teaming under operational constraints, and more \cite{hagenow2024system, aromoye2025significant}. The DARPA Subterranean Challenge demonstrated how multi-robot systems deployed without proper human-in-the-loop supervision and capability-aware coordination can lead to mission failures \cite{agha2021nebula}. Similar transitions toward human-autonomous collaboration are being explored in other confined industrial environments, such as the development of autonomous snow-plowing systems for airports \cite{langen2024classifying}.

From a systems engineering perspective, the challenge of deploying hybrid human-robot surveillance networks is fundamentally one of architecture trade-space exploration. Determining where to locate command centers, what operational level to build them at, and how to staff them with humans and robots constitutes a set of interdependent design decisions spanning the strategic, tactical, and operational layers of system architecture \cite{INCOSE2023}. Current SE practice lacks quantitative methods to evaluate facility and workforce implications of different human-robot teaming configurations during early concept development \cite{amokrane2024combining}. Operations research methods, when embedded within the SE process, can fill this gap by enabling rigorous trade-off analysis across technology maturity scenarios. The subsequent System Overview and Stakeholder Requirements sections formalize this perspective, which we revisit in the Discussion.

Traditionally, the facility location problem (FLP) and services coverage are modeled as covering problems in operations research, in which classical models minimize facility costs while covering demand points, or maximize demands served given available servers \cite{li2011covering}. Yet existing FLP models remain inadequate for modern petroleum security: they treat resources as homogeneous units, ignoring the distinction between autonomous systems capable of continuous patrol and human operators essential for ethical escalation decisions \cite{ye2024human}. Conventional covering models also impose uniform service standards despite stark asymmetries (i.e. protecting a Tier-1 refinery where breach enables catastrophic sabotage differs fundamentally from monitoring remote pipeline segments) \cite{cisa2024plan}. With ransomware attacks on oil and gas surging 935\% between 2024--2025 \cite{zscaler2025ransomware}, the absence of human-in-the-loop constraints creates a dangerous blind spot as organizations deploy increasingly autonomous fleets \cite{adams2024human, pohler2025keeping}.

We address these gaps by proposing a conceptual model bridging the  hierarchical criticality structure of petroleum infrastructure with human-in-the-loop supervision constraints that reflect both regulatory requirements for autonomous system deployment and their maturity in security-critical applications \cite{pohler2025keeping, cisa2024plan}. To that end, we consider: (i) tiered infrastructure criticality with differentiated service level agreements \cite{church2004identifying}; (ii) human-robot co-dispatch with supervision ratio constraints \cite{hagenow2023coordinated}; (iii) redundant coverage calibrated to asset vulnerability \cite{hogan1986concepts}; and (iv) command center capacity constraints \cite{daskin1997network}. We then build a mathematical model formulation \textit{Human-Robot Co-Dispatch Facility Location Problem} (HRCD-FLP) to bridge classical facility location with human-robot teaming for critical infrastructure protection. We demonstrate applicability through a Dhahran district case study across different technology maturity scenarios, showing how the supervision ratio serves as a systems engineering design variable for evaluating deployment architectures. Our overall goal is to study how to integrate rigorous FLPs that has transformed service sectors in the past into security-critical applications considering human-robot teaming trend in practice.

\section{Literature Review}

This work bridges three research domains: critical infrastructure protection (CIP), facility location problems (FLPs), and human-robot collaboration (HRC) for surveillance.

\subsection{Critical Infrastructure Protection}
Petroleum infrastructure protection has received sustained attention since 2001. TSA Pipeline Security Guidelines and DHS frameworks classify facilities based on criticality---target viability, energy supply importance, and weaponization potential \cite{bajpai2007securing, pashchenko2024main}. Church and Scaparra's $r$-interdiction models identify infrastructure elements whose loss maximally degrades system performance, establishing that strategic fortification must account for attacker-defender dynamics \cite{church2004identifying, scaparra2008bilevel}. Contemporary security increasingly leverages autonomous systems; Saudi Aramco has deployed drones for such purposes. However, optimization frameworks for strategic placement of hybrid human-robot dispatch centers remain absent \cite{alsuwailem2022integrated, al2022managing}.

\subsection{Facility Location Problems}
The FLP originates from Weber's industrial location theory and Hakimi's switching center placement \cite{hakimi1964optimum}. Classical methods in these domains include the Set Covering Location Problem (SCLP) or Maximal Covering Location Problem (MCLP). SCLP seeks minimum facilities ensuring demand coverage within service thresholds \cite{toregas1971location}, while the MCLP maximizes covered demand given fixed facilities \cite{church1974maximal, megiddo1983maximum}. The Capacitated Facility Location Problem (CFLP) adds per-facility demand constraints \cite{daskin1997network}. Extensions include backup coverage \cite{hogan1986concepts} and bilevel interdiction-fortification models \cite{scaparra2008bilevel, ghaderi2024bi}. Emergency medical services literature has developed tiered response systems with different vehicle types serving different priorities \cite{farahani2012covering}. These extensions conceptually adds parallel to human-robot teaming, but heterogeneous resource integration with supervision requirements aspects remains unaddressed.

\subsection{Human-Robot Collaboration}
Autonomous security robots now offer 24/7 patrolling with thermal imaging across critical infrastructure in over 15 countries \cite{ye2024human}. Despite capabilities, ethical and regulatory considerations require human oversight for escalation decisions, manifesting as supervision ratio constraints \cite{chen2007human}. The human-robot teaming literature identifies operator multitasking, trust calibration, and cognitive workload as key collaboration factors \cite{chen2007human, haring2013influence}. Recent HSI research has emphasized the importance of functional analysis to clarify task distributions between automated systems and human operators in semi-automated environments to ensure system security \cite{amokrane2024combining}. Operations research has addressed human-robot task allocation in warehouse environments \cite{boysen2019warehousing}, but strategic location for security remains unexamined.

\subsection{Research Gaps}
Three gaps emerge at these domain intersections. First, CIP literature lacks optimization models for hybrid human-robot workforces. Second, FLP literature has not incorporated human-robot constraints including supervision ratios and differential efficiency rates. Third, HRC literature focuses on operational coordination with limited attention to strategic facility location. This paper addresses these gaps through an integrated HRCD-FLP formulation that combines criticality, capacitated facility location, explicit human-robot modeling, supervision constraints, and minimum utilization requirements.

\section{System Overview}
The system-of-interest comprises interacting elements. First, \textit{demand sites} represent petroleum assets requiring surveillance coverage quantified in Surveillance Coverage Units (SCU), encompassing high-criticality concentrated facilities (refineries, processing plants, storage terminals) and distributed linear assets (pipeline segments, valve stations). Second, \textit{command centers} are candidate facility locations from which security resources are dispatched, each characterized by construction costs, operational overhead, and physical capacity constraints. Third, \textit{security resources} constitute a heterogeneous fleet of human personnel and autonomous robotic units with different capabilities, costs, and constraints.

\subsection{Stakeholder Requirements}
The design is driven by five key stakeholder requirements. \textit{Coverage completeness} mandates that all demand sites receive 100\% of required security coverage without exception. \textit{Response time SLAs} require that command centers may only serve demand sites within maximum allowable response distances, with thresholds differentiated by asset criticality tier. \textit{Human-in-the-loop constraints} stipulate that autonomous systems must operate under human supervision at prescribed ratios (e.g., 1:5 human-to-robot) to ensure ethical decision-making capacity for threat escalation scenarios. \textit{Economic efficiency} requires minimizing total infrastructure and operational costs subject to coverage and supervision constraints. Finally, \textit{infrastructure utilization thresholds} ensure that opened facilities achieve minimum occupancy rates to justify capital investment and prevent proliferation of underused command centers.

\begin{figure*}[t]
\centering
\includegraphics[width=1.0\linewidth]{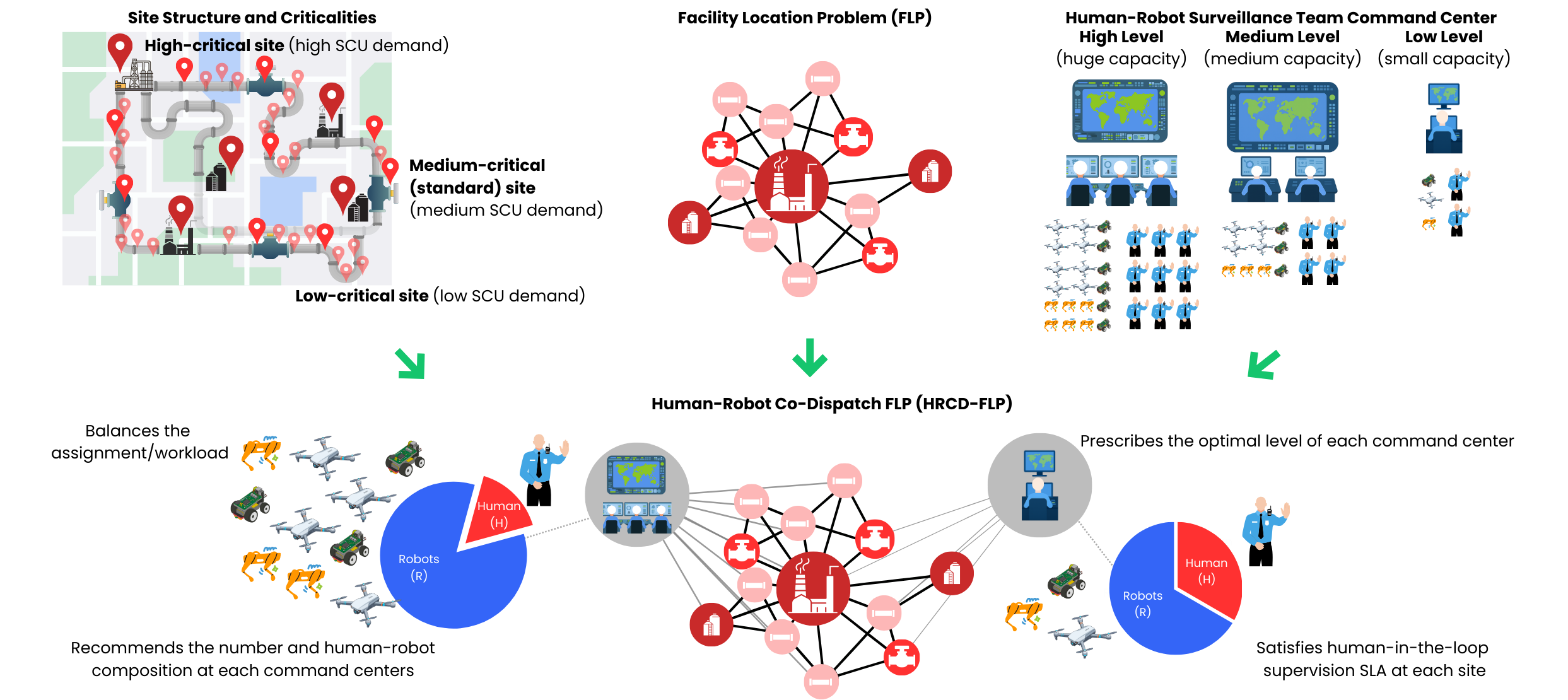}
\caption{Conceptual architecture of the HRCD-FLP framework.}
\label{fig:conceptual_model}
\end{figure*}

\subsection{System Architecture}
The key aspects of the model includes heterogeneous resources, encompassing three hierarchical decision layers. At the \textit{strategic layer}, the model determines optimal locations among candidate sites and selects operational levels (High, Medium, Low) for each command center, where higher levels provide greater capacity and faster response capabilities at increased fixed cost. The \textit{tactical layer} assigns each demand site to exactly one command center through a single-sourcing constraint, ensuring unified command responsibility and SLA compliance while simplifying operational coordination. At the \textit{operational layer}, the model determines the resource mix of humans and robots at each facility, satisfying aggregate coverage requirements, physical capacity limits, and supervision ratio constraints.

The proposed system design involves fundamental trade-offs. The coverage-cost trade-off reflects tension between deploying more facilities (ensuring tighter coverage but increasing fixed costs) versus fewer facilities (reducing capital expenditure but risking SLA violations for remote assets). The human-robot resource trade-off balances robots' higher efficiency (3--5$\times$ human patrol capacity) and lower marginal cost against their requirement for human supervision and inability to exercise judgment for threat escalation. Facility level selection weighs premium construction and operational costs of higher-level facilities against their faster response times and greater capacity. Finally, the centralization-distribution trade-off considers whether consolidated high-capacity centers achieving economies of scale outperform distributed lower-capacity outposts that reduce response distances to remote assets. \autoref{fig:conceptual_model} illustrates the system and the proposed modeling framework. The following section presents the formal mathematical specification.

\section{Model Formulation}

We formulate the problem as a multi-level capacitated facility location model with heterogeneous resources, capturing the key trade-offs identified in the preceding system analysis.

\subsection{Problem Setting}
Consider a set of candidate command center (CC) locations $I = \{1, \ldots, m\}$ and a set of demand sites $J = \{1, \ldots, n\}$ requiring security surveillance. Each candidate location $i \in I$ may be developed into a command center. Demand sites require security coverage measured in Surveillance Coverage Units (SCU), a composite metric capturing patrol frequency, sensor density, and response readiness appropriate to asset criticality. This demand is satisfied by deploying two resource types $k \in K = \{\text{Robot}, \text{Human}\}$ from assigned CCs. Each CC is built at one of three operational levels $l \in L = \{\text{High}, \text{Medium}, \text{Low}\}$, where each level is characterized by distinct cost structures, resource capacities, and response time capabilities reflecting the CC infrastructure quality and technological sophistication. Robots provide high-efficiency continuous patrol capability, while humans contribute supervisory oversight and contextual judgment for threat escalation. The resource mix at each facility is governed by both site-specific coverage requirements and global supervision policies ensuring human-in-the-loop control.

\subsection{Parameters}
The model is parameterized by cost, capacity, and demand data. For costs, let $F_{il}$ denote the fixed construction and operational overhead for establishing facility $i$ at level $l$, and let $C_{ik}$ represent the unit deployment cost for resource type $k$ at location $i$, capturing salary and benefits for human personnel or amortized capital and maintenance costs for robotic units.

Capacity parameters include $\text{MAXCAP}_{lk}$, the maximum number of resource type $k$ that a level-$l$ facility can accommodate, and $\text{MINCAP}_{lk}$, the minimum resource deployment required to maintain operational readiness if a facility is opened. Response capability is captured by $t_{ijl}$, the response time from facility $i$ at level $l$ to demand site $j$, where higher-level facilities may deploy faster assets (e.g., VTOL drones versus standard quadcopters). Each demand site $j$ has an associated service level agreement $S_j$ specifying maximum allowable response time, differentiated by asset criticality tier.

Demand parameters capture the surveillance workload and required resource composition at each site. We introduce the \textit{Surveillance Coverage Unit} (SCU) as a composite demand metric that quantifies the total security workload required at a protected asset. SCU captures three operational dimensions: \textit{patrol frequency}, the number of inspection passes required per shift to maintain situational awareness \cite{adams2024human, ye2024human}; \textit{sensor density}, the number of concurrent monitoring feeds needed for continuous surveillance \cite{gouglidis2018surveillance}; and \textit{response readiness}, the standby posture level prescribing how rapidly dispatch-capable units must be available to respond to detected anomalies \cite{li2011covering, farahani2012covering}. Higher values across these dimensions yield higher SCU requirements. In the petroleum context studied here, SCU values are calibrated to infrastructure criticality tier \cite{cisa2024plan, church2004identifying}: Tier-1 sites require 15--21 SCU, Tier-2 standard sites 8--15 SCU, and Tier-3 sites 3--8 SCU.

The site-specific human-robot mix ratio $\alpha_j$ reflects task complexity and escalation likelihood. For Tier-1 sites, $\alpha_j \in [1.01, 2.0]$ (human-dominant); Tier-2 sites, $\alpha_j = 1.0$ (equal split); Tier-3 sites, $\alpha_j \in [0.1, 0.99]$ (robot-dominant). The global supervision ratio $\alpha$ encodes the minimum human-to-robot ratio mandated by policy.

\subsection{Decision Variables}
The model determines three classes of decisions corresponding to the strategic, tactical, and operational layers identified in the system architecture. The binary variable $x_{il} \in \{0, 1\}$ equals 1 if candidate location $i$ is developed at operational level $l$, and 0 otherwise, representing the strategic facility location and level selection. The binary variable $y_{ij} \in \{0, 1\}$ equals 1 if demand site $j$ is assigned to facility $i$ for service, encoding the tactical demand-to-facility assignment. Finally, the integer variable $z_{ik} \in \mathbb{Z}^+$ specifies the number of resource type $k$ deployed at facility $i$, determining the operational resource mix.

\subsection{Objective Function}
The objective minimizes total system cost, comprising fixed infrastructure costs and variable resource deployment costs:
\begin{equation}
\min_{x, y, z} \quad Z = \sum_{i \in I} \sum_{l \in L} F_{il} \, x_{il} + \sum_{i \in I} \sum_{k \in K} C_{ik} \, z_{ik}.
\label{eq:objective}
\end{equation}
\noindent The first term aggregates construction and operational overhead across all opened facilities, while the second term captures ongoing personnel and equipment costs scaled by deployment quantities.

\subsection{Constraints}
The optimization is subject to constraints encoding stakeholder requirements and operational feasibility.

\subsubsection{Facility Configuration} Each candidate location may host at most one facility, and if developed, must be assigned exactly one operational level:
\begin{equation}
\sum_{l \in L} x_{il} \leq 1, \quad \forall i \in I.
\label{eq:single_level}
\end{equation}

\subsubsection{Demand Coverage} Every demand site must be assigned to at least one command center:
\begin{equation}
\sum_{i \in I} y_{ij} \geq 1, \quad \forall j \in J.
\label{eq:demand_assignment}
\end{equation}
\noindent While the inequality permits redundant assignment for backup coverage at critical sites, practical implementations typically achieve single-sourcing to ensure unified command responsibility.

\subsubsection{Assignment Feasibility} A demand site may only be assigned to an active facility, which linking tactical assignments to strategic location decisions:
\begin{equation}
y_{ij} \leq \sum_{l \in L} x_{il}, \quad \forall i \in I, \;
\forall j \in J.
\label{eq:logical_link}
\end{equation}

\subsubsection{Service Level Compliance} Assignments must respect response time constraints, where a facility at level $l$ may serve site $j$ only if the response time falls within the SLA threshold. Using big-$M$ formulation, we have
\begin{equation}
t_{ijl} \, x_{il} \leq S_j + M(1 - y_{ij}), \quad \forall i \in I, \;
\forall j \in J, \; \forall l \in L.
\label{eq:sla}
\end{equation}
\noindent This constraint ensures that high-criticality Tier-1 assets with stringent SLAs can only be served by nearby facilities or those equipped with rapid-deployment capabilities.

\subsubsection{Resource Capacity Bounds} Resource deployment at each facility must respect physical capacity limits determined by the selected operational level:
\begin{equation}
z_{ik} \leq \sum_{l \in L} \text{MAXCAP}_{lk} \, x_{il}, \quad \forall i \in I, \;
\forall k \in K.
\label{eq:max_cap}
\end{equation}
\noindent The upper bound prevents over-allocation beyond infrastructure capacity. The lower bound ensures minimum staffing for operational readiness when a facility is opened, preventing proliferation of skeleton crews:
\begin{equation}
z_{ik} \geq \sum_{l \in L} \text{MINCAP}_{lk} \, x_{il}, \quad \forall i \in I, \;
\forall k \in K.
\label{eq:min_cap}
\end{equation}

\subsubsection{Coverage Satisfaction} Deployed resources must be sufficient to meet the aggregate surveillance demand of all assigned sites, accounting for the site-specific human-robot task allocation. Robot deployment must satisfy:
\begin{equation}
z_{i,\text{Robot}} \geq \sum_{j \in J} \frac{D_j}{1 + \alpha_j} \, y_{ij}, \quad \forall i \in I.
\label{eq:robot_coverage}
\end{equation}
\noindent Meanwhile, human deployment must satisfy:
\begin{equation}
z_{i,\text{Human}} \geq \sum_{j \in J} \frac{D_j \, \alpha_j}{1 + \alpha_j} \, y_{ij}, \quad \forall i \in I .
\label{eq:human_coverage}
\end{equation}
\noindent Here, $\alpha_j$ governs the demand split: sites with higher $\alpha_j$ require proportionally more human involvement due to task complexity or escalation sensitivity.

\subsubsection{Human-in-the-Loop Supervision}
A global policy constraint ensures adequate human oversight for autonomous operations, requiring the human contingent to scale with robot fleet size:
\begin{equation}
z_{i,\text{Human}} \geq \alpha \cdot z_{i,\text{Robot}}, \quad \forall i \in I.
\label{eq:supervision}
\end{equation}
\noindent This constraint operationalizes the human-in-the-loop requirement central to this work. As organizations deploy larger autonomous fleets, the supervision ratio $\alpha$ (e.g., 1:5 indicating one human per five robots) mandates proportional human presence for ethical decision-making and regulatory compliance. The ratio $\alpha$ serves as a policy lever reflecting technology maturity: conservative deployments with less proven autonomy require higher ratios, while mature systems with established trust may operate under relaxed supervision.

\subsection{Model Characteristics and Heuristics}
The HRCD-FLP formulation constitutes a mixed-integer linear program (MILP) with $|I| \cdot |L| + |I| \cdot |J|$ binary variables and $|I| \cdot |K|$ integer variables. The problem generalizes the classical capacitated facility location problem through three extensions: multi-level facility selection introducing discrete infrastructure tiers, heterogeneous resources with differential efficiency and supervision requirements, and the supervision ratio constraint \eqref{eq:supervision} coupling human and robot deployment decisions. These extensions increase model expressiveness at the cost of computational complexity, motivating the solution approach presented in the following section.

The optimization problem posed by the HRCD-FLP is NP-hard, involving complex interdependencies between facility location, level selection, and resource allocation. To address this computational challenge, we employed a hybrid solution strategy that balances optimality with scalability. For smaller instances and model validation, we used an exact solution approach based on efficient branch-and-bound implemented via the Gurobi solver. This exact solver serves as a baseline for benchmarking performance. Recognizing the computational limits of exact methods for large-scale deployments (as will be shown in one of our experiments), we further developed a specialized two-stage metaheuristic.

The first stage consists of a multi-level constructive greedy algorithm that iteratively assigns demand sites to facilities by minimizing marginal costs while dynamically adjusting facility levels to meet SLA and capacity constraints. The second stage refines this initial solution through a "Best Improvement" local search, which explores the solution space using shift, swap, drop, and open moves to escape local optima and enhance solution quality. The overall procedure is outlined in Algorithm \ref{alg:heuristic}.

\begin{algorithm}[t]
\caption{Two-Stage Heuristic for HRCD-FLP}
\label{alg:heuristic}
\footnotesize
\begin{algorithmic}[1]
\State \textbf{Input:} Sites $I$, Demand $J$, Levels $L$, Params (Costs, Demand, SLA, etc.)
\State \textbf{Output:} Facility locations $x$, Assignments $y$, Resources $z$
\Statex
\State \textit{\textbf{Stage 1: Constructive Greedy}}
\For{$j \in J$}
    \State $min\_cost \gets \infty$, $best\_i \gets \text{null}$
    \For{$i \in I$}
        \If{facility $i$ can serve $j$ (SLA \& feasible level)}
            \State $\Delta C \gets$ Cost to add $j$ to $i$ (incl. level upgrades)
       
     \If{$\Delta C < min\_cost$}
                \State $min\_cost \gets \Delta C$, $best\_i \gets i$
            \EndIf
        \EndIf
    \EndFor
    \State Assign $j \to best\_i$;
    \State Update $x_{best\_i}$ to min feasible level
    \State Update resources $z_{best\_i}$ based on assigned demand
\EndFor
\Statex
\State \textit{\textbf{Stage 2: Local Search}}
\Repeat
    \State $improved \gets \textbf{false}$
    \State \textit{// Moves use random sampling for large neighborhoods}
    \If{\textsc{ShiftMove}$()$: move $j$ to different $i$ saves cost}
         \State Apply move;
         \State $improved \gets \textbf{true}$
    \ElsIf{\textsc{SwapMove}$()$: swap $j_1, j_2$ between $i_1, i_2$ saves cost}
         \State Apply move;
         \State $improved \gets \textbf{true}$
    \ElsIf{\textsc{DropMove}$()$: close $i$, redistribute sites saves cost}
         \State Apply move;
         \State $improved \gets \textbf{true}$
    \ElsIf{\textsc{OpenMove}$()$: open new $i$ nearby saves cost}
         \State Apply move;
         \State $improved \gets \textbf{true}$
    \EndIf
\Until{no improvement for $N$ consecutive iterations}
\end{algorithmic}
\end{algorithm}

\section{Numerical Experiments}
The study uses a hybrid dataset designed to simulate one of the largest petroleum site complexes in Saudi Arabia, combining real-world geospatial topology with synthetic parameter based on industrial security and state-of-the-art human-robot teaming standards. We conducted two sets of experiments. The code is available at {\color{blue}{\url{https://github.com/ai-vnv/Petro-HRCD-FLP}}}.

The first experimental environment comprises 15 candidate locations and 50 demand sites, distributed to mimic pipeline corridors and scattered high-value assets. Geodesic distances were calculated to ensure realistic response time modeling. 
The 50 demand sites are distributed across three criticality tiers: approximately 10\% Tier-1, 25\% Tier-2, and 65\% Tier-3, yielding a total demand of approximately 430 SCU with Tier-1 sites contributing roughly 35\% of total SCU despite comprising only 10\% of sites. This demand concentration, combined with Tier-1's stringent 5-minute SLA, anchors command center placement near critical nodes while dispersed Tier-3 sites allow greater assignment flexibility.
Facility capabilities were modeled across three tiers: High-level facilities represent premium infrastructure with 1.5x base fixed costs but high capacity ($\approx 240$ robots) and rapid response assets; Medium-level facilities reflect standard infrastructure; and Low-level facilities act as satellite outposts with lower costs but limited capacity and slower response times. This problem represents a small-size problem where both methods terminate and obtain solutions.

To evaluate the impact of technological maturity, we defined three scenarios: Conservative, Balanced, and Future. These scenarios vary the global supervision ratio ($\alpha$), the robot cost multiplier, and the site-specific mix scaler. The Conservative scenario represents current operational constraints with a 1:3 human-to-robot supervision ratio and baseline robot costs. The Balanced scenario reflects near-term improvements with a 1:5 ratio and 10\% lower robot costs. The Future scenario projects a mature autonomous ecosystem with a 1:10 ratio and 20\% cost reduction, allowing for highly automated operations. To solve both cases, we set $N=100$ for Algorithm \ref{alg:heuristic}.

The results for this small-scale problem are summarized in \autoref{tab:experiment_results}, with the distribution of CCs shown in \autoref{fig:cc_levels}. Meanwhile, the comparison of the facility resources obtained by the exact and heuristic methods are summarized in \autoref{fig:res_exact} and \autoref{fig:res_heuristic}, respectively. The breakdown for each conservative, balanced, and future maturity scenario is shown in \autoref{fig:result_conservative}, \autoref{fig:result_balanced}, and \autoref{fig:result_future}, respectively.

The second experimental environment simulates larger-scale deployment of the model, dealing with 500 command center candidates and 5,000 sites demanding varying SCUs and SLAs. The results for running both the exact method (using Gurobi with 3,600 seconds time limit and 1\% optimality gap) vs. heuristics is summarized in Table \ref{tab:exp_results_large}.

\section{Discussion}

\begin{table*}[!t]
\centering
\caption{Experiment Results for Small-size Problem (15 CC candidates, 50 sites)}
\label{tab:experiment_results}
\begin{tabular*}{\textwidth}{@{\extracolsep{\fill}}llcccccc@{}}
\toprule
\textbf{Scenario} & \textbf{Method} & \textbf{Facilities} & \textbf{Robots} & \textbf{Humans} & \textbf{Cost (\$)} & \textbf{Time (s)} & \textbf{Gap (\%)} \\
\midrule
Conservative & Exact & 4 & 207 & 149 & 692,450 & 1.32 & -- \\
 & Heuristic & 4 & 207 & 149 & 692,450 & 6.61 & 0.00 \\
\addlinespace
Balanced & Exact & 4 & 229 & 127 & 610,175 & 0.24 & -- \\
 & Heuristic & 4 & 229 & 127 & 620,175 & 4.46 & 1.64 \\
\addlinespace
Future & Exact & 3 & 256 & 98 & 508,000 & 0.25 & -- \\
 & Heuristic & 4 & 258 & 98 & 529,200 & 6.94 & 4.17 \\
\bottomrule
\end{tabular*}
\end{table*}

\begin{table*}[!t]
\centering
\caption{Experiment Results for Large-size Problem (500 CC candidates, 5,000 sites)}
\label{tab:exp_results_large}
\begin{tabular*}{\textwidth}{@{\extracolsep{\fill}}llcccccc@{}}
\toprule
\textbf{Scenario} & \textbf{Method} & \textbf{Facilities} & \textbf{Robots} & \textbf{Humans} & \textbf{Cost (\$)} & \textbf{Time (s)} & \textbf{Gap (\%)} \\
\midrule
Conservative & Exact* & -- & -- & -- & -- & -- & -- \\
 & Heuristic & 500 & 21,528 & 17,830 & -- & 143.06 & -- \\
\addlinespace
Balanced & Exact* & -- & -- & -- & -- & -- & -- \\
 & Heuristic & 500 & 24,136 & 15,371 & 78,141,930 & 28.02 & -- \\
\addlinespace
Future & Exact & 306 & 27,368 & 11,945 & 59,951,160 & 2,467.59 & -- \\
 & Heuristic & 500 & 27,462 & 12,045 & 68,486,460 & 27.05 & 14.24 \\
\bottomrule
\multicolumn{8}{@{}l}{\footnotesize $^*$Method failed to terminate or returned infeasible at 3,600s time limit.}
\end{tabular*}
\end{table*}

We discuss the results of the optimization model across the scenarios, particularly analyzing the optimization performance, resource allocations, and demand assignments.

\subsection{Optimization Performance}
The experimental evaluation demonstrates the efficacy of the proposed model across the three defined technological scenarios. As summarized in Table \ref{tab:experiment_results}, the exact method consistently identified optimal solutions with lower costs compared to the heuristic approach, although the Heuristic demonstrated competitive performance with optimality gaps averaging under 2\% for small-scale problems. Notably, the exact solver proved highly efficient for small test instances, with solution times under 1.5 seconds; however, the heuristic required approximately 5--7 seconds due to its iterative search process. This computational overhead represents a key trade-off: \textbf{for small-scale problems, the exact method dominates in both solution quality and speed}, making the heuristic less attractive when problem size permits exact optimization.

However, Table \ref{tab:exp_results_large} reveals a dramatic reversal at scale. For the large-scale problem (500 candidates, 5,000 sites), the exact method failed to terminate within the 3,600-second time limit for Conservative and Balanced scenarios, and required over 41 minutes for the Future scenario. In contrast, \textbf{the heuristic produced feasible solutions in under 3 minutes across all scenarios}---achieving a 91$\times$ speedup in the Future scenario where comparison is possible. This scalability comes at a cost: the heuristic's optimality gap reached 14\% for large problems, and it selected more facilities (500 vs.\ 306 in the Future scenario), indicating suboptimal consolidation. These results underscore the heuristic's value as a practical tool for large-scale deployments where exact methods become intractable, while acknowledging its limitations in solution quality.
We note that the current evaluation relies on synthetic operational 
parameters calibrated from published standards rather than proprietary 
data, a limitation common in facility location literature where access 
to sensitive infrastructure data is restricted. The heuristic's tendency 
toward facility proliferation likely stems from the greedy construction 
stage, which opens new facilities rather than evaluating whether 
upgrading an existing facility's level would be more cost-effective. 
Validation against real deployments and alternative network topologies 
remain important future directions.

\subsection{Resource Allocation Analysis}
\begin{figure*}[p]
    \centering
    \includegraphics[width=0.9\linewidth]{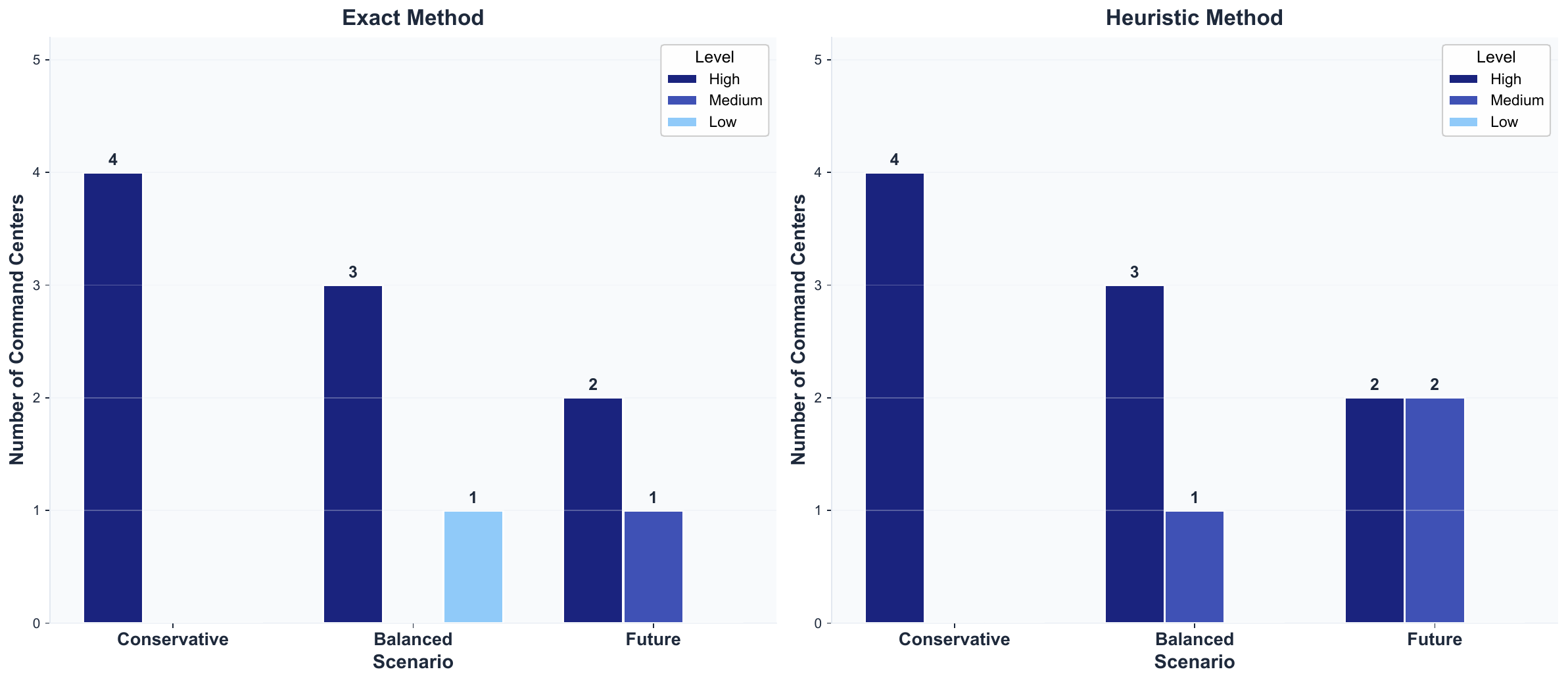}
    \caption{Distribution of Command Center Levels across Scenarios}
    \label{fig:cc_levels}
\end{figure*}

Analyzing the resource allocation reveals distinct trends driven by the supervision constraints and cost parameters. As illustrated in Figure \ref{fig:cc_levels}, the distribution of command center levels shifts across scenarios. In the Conservative scenario, the system favors a distributed network of High-level facilities to accommodate the large human workforce required by the 1:3 supervision ratio. Conversely, as the supervision ratio relaxes in the Future scenario, the model consolidates operations into a leaner network of facilities, reflecting the improved economics of automation.

\begin{figure*}[p]
    \centering
    \includegraphics[width=0.9\linewidth]{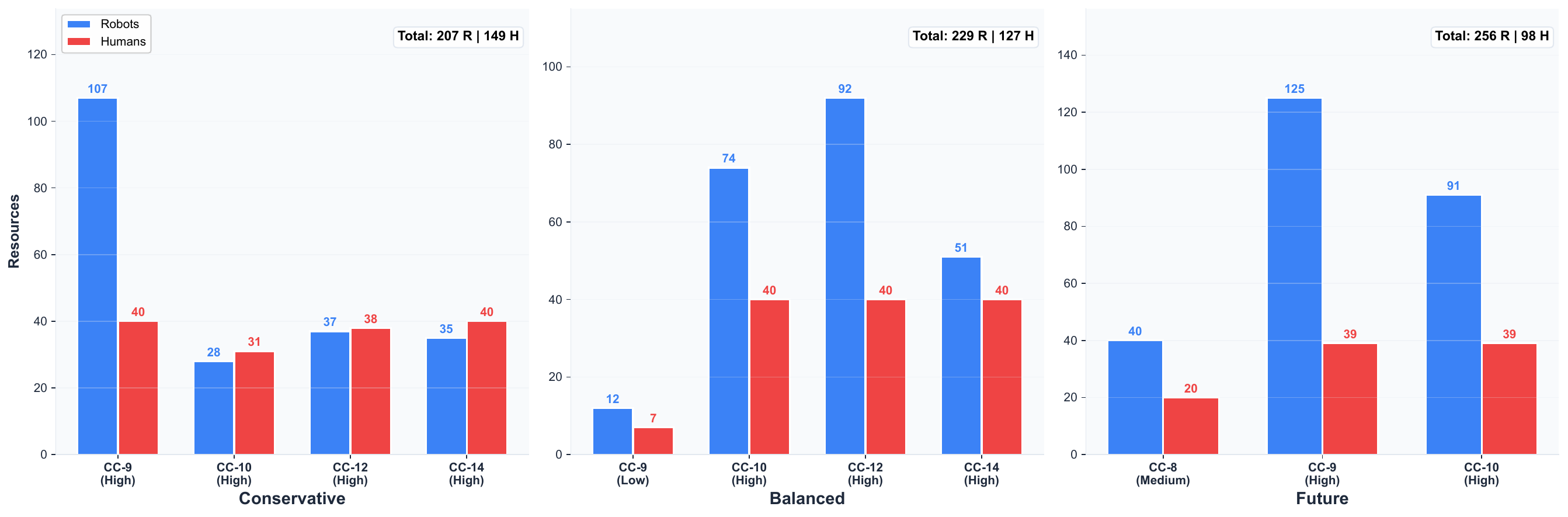}
    \caption{Facility Resource Allocation (Exact Method)}
    \label{fig:res_exact}
\end{figure*}

\begin{figure*}[p]
    \centering
    \includegraphics[width=0.9\linewidth]{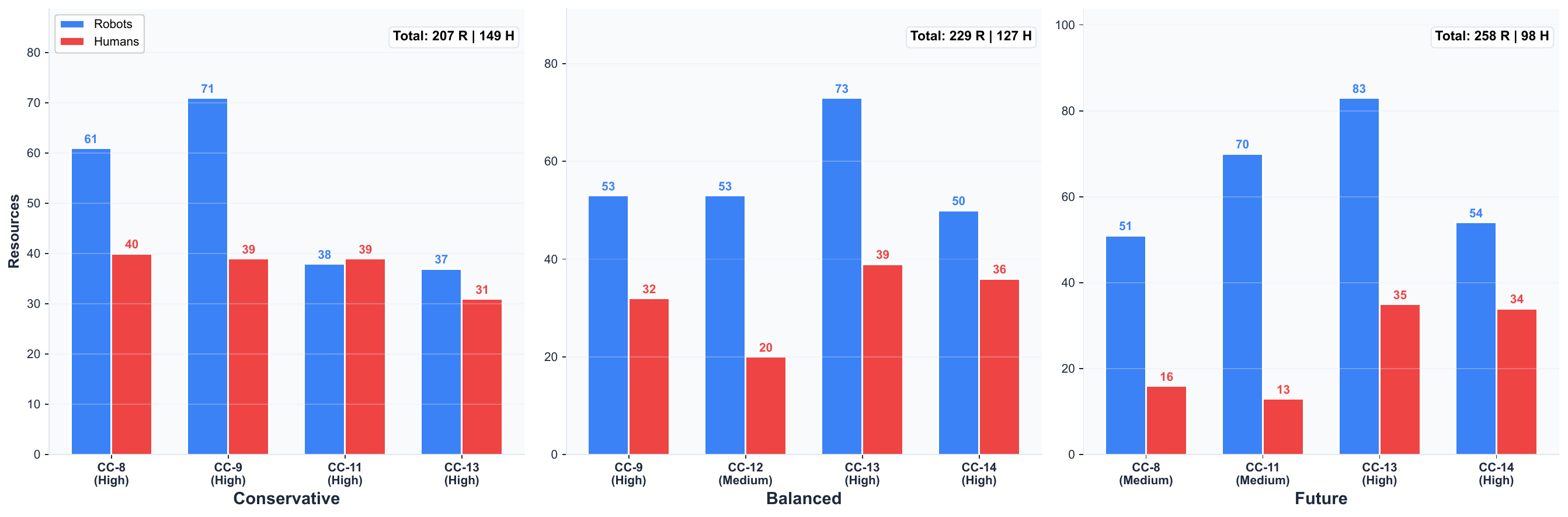}
    \caption{Facility Resource Allocation (Heuristic Method)}
    \label{fig:res_heuristic}
\end{figure*}

Figures \ref{fig:res_exact} and \ref{fig:res_heuristic} detail the specific resource counts for the Exact and Heuristic solutions, respectively. For small-scale problems, the methods achieve comparable resource allocations---in the Conservative scenario, both methods converged to identical solutions (207 robots, 149 humans). The Exact method (Figure \ref{fig:res_exact}) achieves marginally tighter alignment of resources to demand in Balanced and Future scenarios. However, the efficiency gap becomes pronounced at scale: in the large-scale Future scenario, the exact method opens only 306 facilities compared to the heuristic's 500---a 1.6$\times$ difference reflecting the heuristic's greedy tendency to open more facilities rather than consolidate into optimal high-capacity centers. This facility proliferation, while ensuring feasibility, results in the 14\% cost premium observed in Table \ref{tab:exp_results_large}.

\begin{figure*}[p]
    \centering
    \includegraphics[width=0.8\linewidth]{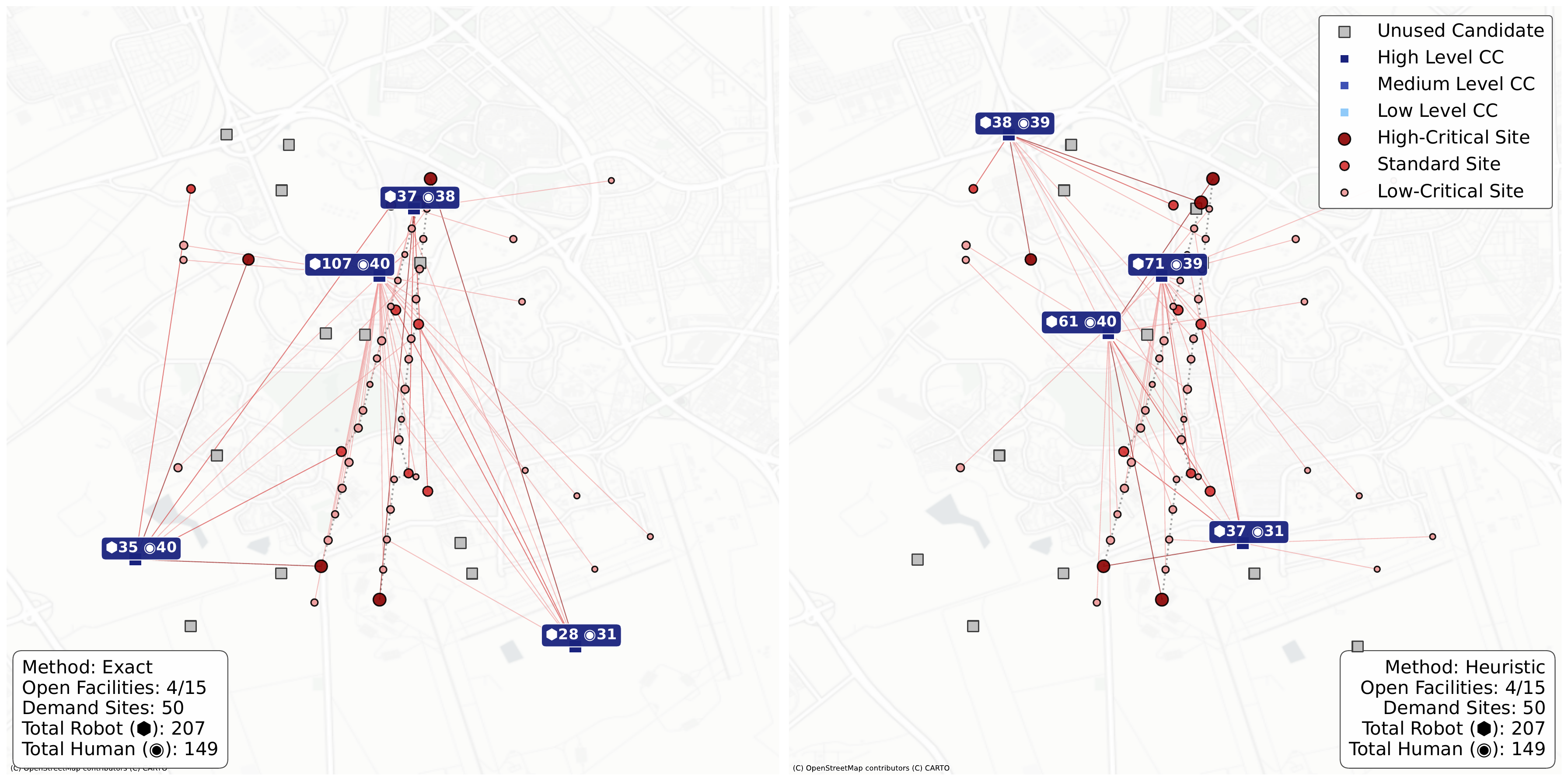}
    \caption{Conservative Scenario Breakdown}
    \label{fig:result_conservative}
\end{figure*}

\begin{figure*}[p]
    \centering
    \includegraphics[width=0.8\linewidth]{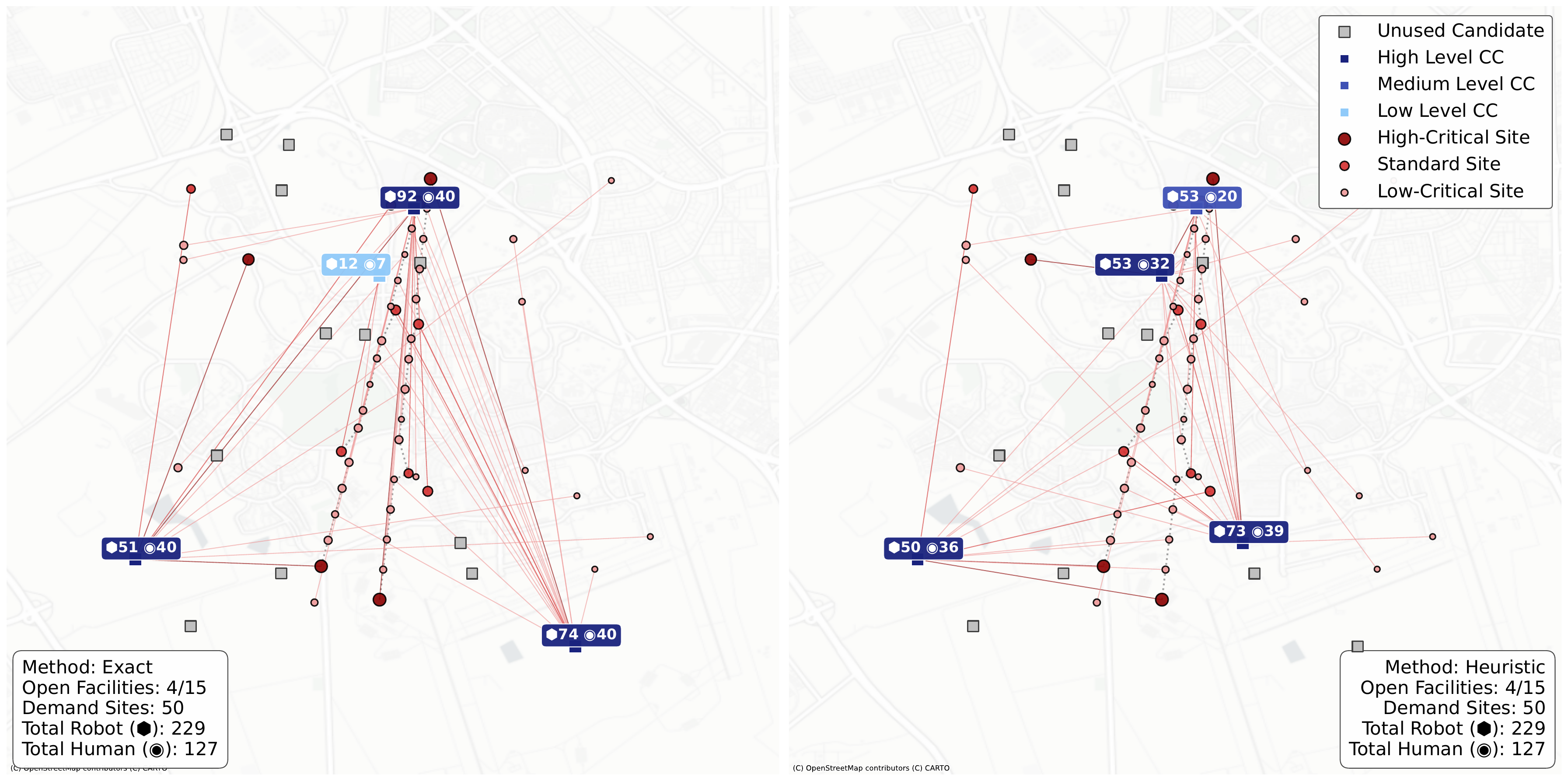}
    \caption{Balanced Scenario Breakdown}
    \label{fig:result_balanced}
\end{figure*}

\begin{figure*}[p]
    \centering
    \includegraphics[width=0.8\linewidth]{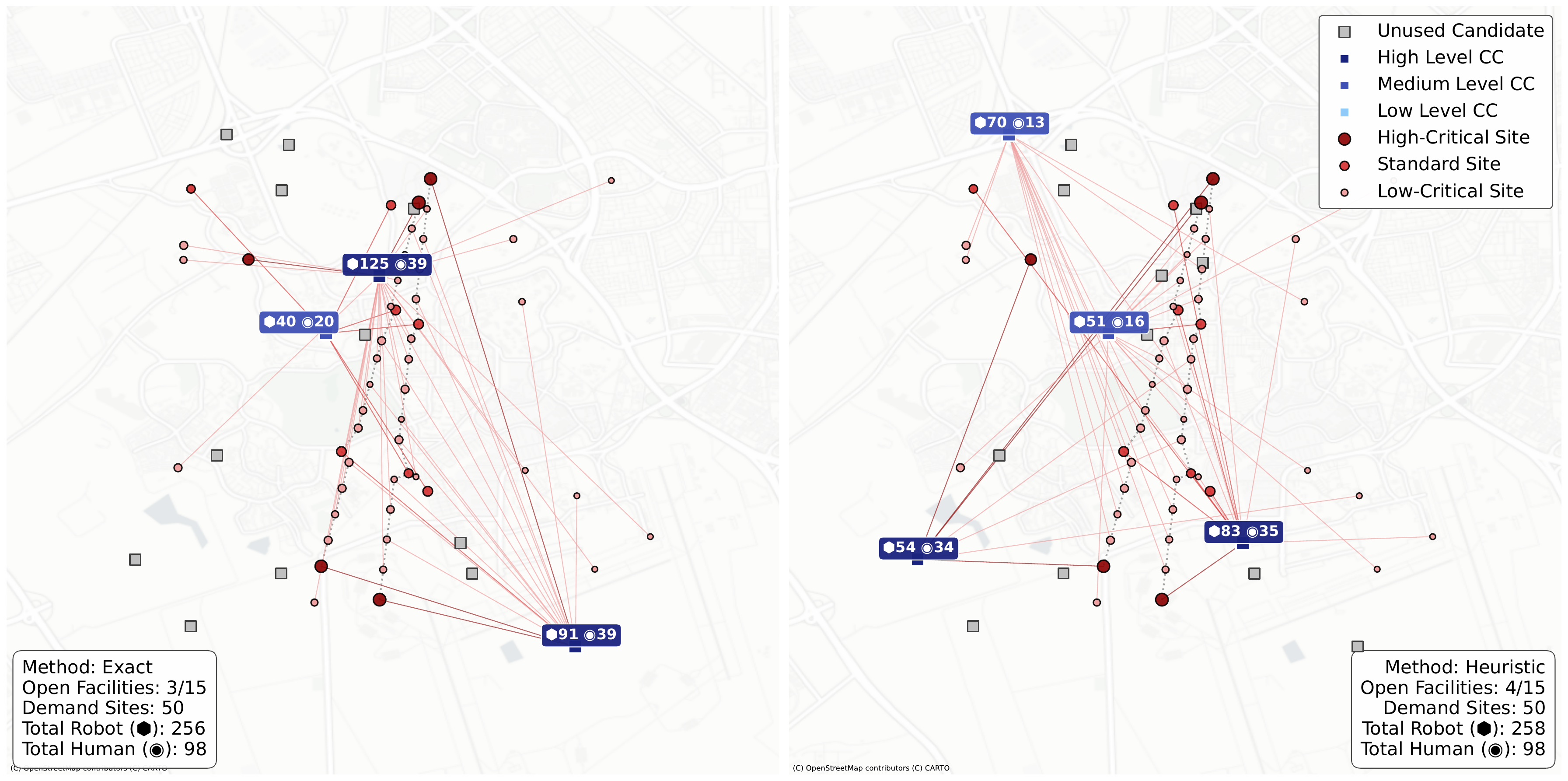}
    \caption{Future Scenario Breakdown}
    \label{fig:result_future}
\end{figure*}

\subsection{Demand Assignment Analysis}
The spatial distribution of demand assignments, as visualized in Figures \ref{fig:result_conservative}, \ref{fig:result_balanced}, and \ref{fig:result_future}, provides further insight into the network topology. The demand assignment analysis shows how the model adapts to the changing constraints. In the Conservative scenario, the dense network of facilities ensures that no demand site is too far from a command center, critical for maintaining the high supervision ratios. As we move to the Balanced and Future scenarios, the "catchment areas" of each facility expand, taking advantage of the reduced need for human proximity and the high efficiency of the autonomous units. This allows a more streamlined security architecture.

\subsection{Systems Engineering Implications}
The HRCD-FLP framework serves as a quantitative tool for architecture trade-space exploration during early concept development. By parameterizing the supervision ratio $\alpha$ as a design variable, the model enables systems engineers to evaluate how changes in autonomy maturity propagate through facility investment, workforce composition, and service level compliance. For example, the transition from Conservative ($\alpha = 1:3$) to Future ($\alpha = 1:10$) not only reduces cost by 27\% but shifts the network from four distributed High-level facilities to three consolidated facilities, a qualitative architecture change emerging from quantitative optimization. The supervision ratio thus functions as a technology readiness policy lever: organizations can use the model to determine what facility investments a given supervision policy requires, or at what ratio network consolidation becomes feasible.

We acknowledge that representing human-robot interaction through constraint ratios is a deliberate simplification. Operational teaming involves richer dynamics including cognitive workload \cite{chen2007human}, trust calibration \cite{haring2013influence}, and degraded autonomy modes. We view the ratio-based formulation as an appropriate first-order approximation for strategic planning, with richer interaction models constituting a natural extension for operational-level refinement.

The HRCD-FLP structure generalizes beyond petroleum to any domain combining distributed assets of varying criticality with mixed human-autonomous teams under supervision constraints, including warehouse logistics \cite{boysen2019warehousing}, emergency medical services \cite{farahani2012covering}, defense, and smart city infrastructure.

Comparing the optimized solutions against a baseline of full human deployment reveals substantial savings. The Conservative scenario yields an estimated 25\% reduction in costs by introducing robotic augmentation. As technology matures, the Balanced and Future scenarios offer dramatic potential savings of 45\% and 65\% respectively. These projections highlight the transformative economic potential of integrating autonomous systems into security operations, provided that the requisite regulatory frameworks for reduced supervision are in place.

\section{Conclusion}
Securing petroleum infrastructure increasingly relies on hybrid human-robot teams, yet strategic planning frameworks have lagged behind operational deployments, leaving organizations without principled methods to determine where to locate command centers and how to staff them. This paper addressed that gap by formulating the Human-Robot Co-Dispatch Facility Location Problem (HRCD-FLP), integrating tiered asset criticality, service level agreements, and human-robot supervision constraints into a unified optimization framework. We found that relaxing supervision ratios from 1:3 to 1:10 enables consolidation from distributed networks to centralized high-capacity centers with up to 27\% cost reduction, which could inform deployment plans. We also showed that exact methods dominate for small instances but become intractable at scale, where heuristics provide feasible solutions within 3 minutes with a 14\% optimality gap. Finally, facility proliferation remains a heuristic inefficiency, opening 1.6$\times$ more facilities than optimal in the comparable scenario. 
From a systems engineering perspective, HRCD-FLP demonstrates how operations research methods can be embedded within the systems engineering process to support architecture decisions for human-robot systems, linking stakeholder requirements to verifiable design constraints through mathematical optimization. While demonstrated in petroleum security, the framework generalizes to any domain combining distributed assets of varying criticality with mixed human-autonomous teams under supervision constraints. Future work should pursue validation against real operational data, enrichment of the supervision model with cognitive workload and trust dynamics, and extension toward dynamic dispatch routing connecting strategic placement to operational coordination.


\bibliographystyle{IEEEtran}
\bibliography{references}

\end{document}